# Automatic Detection of Omissions in Translations

I. Dan Melamed
Department of Computer and Information Science
University of Pennsylvania
Philadelphia, PA, 19104, U.S.A.
melamed@unagi.cis.upenn.edu

## Abstract

ADOMIT is an algorithm for Automatic Detection of OMIssions in Translations. The algorithm relies solely on geometric analysis of bitext maps and uses no linguistic information. This property allows it to deal equally well with omissions that do not correspond to linguistic units, such as might result from word-processing mishaps. ADOMIT has proven itself by discovering many errors in a hand-constructed gold standard for evaluating bitext mapping algorithms. Quantitative evaluation on simulated omissions showed that, even with today's poor bitext mapping technology, ADOMIT is a valuable quality control tool for translators and translation bureaus.

## 1 Introduction

Omissions in translations arise in several ways. A tired translator can accidentally skip a sentence or a paragraph in a large text. Pressing a wrong key can cause a word processing system to delete several lines without warning. Such anomalies can usually be detected by careful proof-reading. However, price competition is forcing translation bureaus to cut down on this labor-intensive practice. An automatic method of detecting omissions can be a great help in maintaining translation quality.

ADOMIT is an algorithm for Automatic Detection of OMIssions in Translations. ADOMIT rests on principles of geometry, and uses no linguistic information. This property allows it to deal equally well with omissions that do not correspond to linguistic units, such as might result from word-processing mishaps. ADOMIT is limited only by the quality of the available bitext map.

The paper begins by describing the geometric properties of bitext maps. These properties enable the Basic Method for detecting omissions. Section 5 suggests how the omission detection technique can be embodied in a translators' tool. The main obstacle to perfect omission detection is noise in bitext maps, which is characterized in Section 6. ADOMIT is a more robust variation of the Basic Method. Section 7 explains how ADOMIT filters out some of the noise in bitext maps. Section 8 demonstrates ADOMIT's performance and its value as a quality control tool.

## 2 Bitext Maps

Any algorithm for detecting omissions in a translation must use a process of elimination: It must first decide which segments of the original text *have* corresponding segments in the translation. This decision requires a detailed description of the correspondence between units of the original text and units of the translation. The original text and its translation constitute a **bitext** (Harris, 1988). A description of the correspondence between the two halves of the bitext is called a **bitext map**. At least two methods for finding bitext maps have been described in the literature (Church, 1993; Melamed, 1996). Both methods output a sequence of corresponding character positions in the two texts. The novelty of the omission detection method presented in this paper lies in analyzing these correspondence points geometrically.

A text and its translation can form the axes of a rectangular **bitext space**, as in Figure 1. The height and width of the rectangle correspond to the lengths of the two texts, in characters. The lower left corner of the rectangle represents the texts' beginnings. The upper right corner represents the texts' ends. If we know other corresponding character positions between the two texts, we can plot them as points in the bitext space. A **bitext map** is the real-valued function obtained by interpolating successive points in a bitext space. The bitext map between two texts that are translations of each other (**mutual translations**) will be injective (one to one).

Bitext maps have another property that is crucial for detecting omissions in translations. There is a very high correlation between the lengths of mutual translations ($\rho = .991$) (Gale & Church, 1991). This implies that the slope of segments of the bitext map function fluctuates very little. The slope of any segment of the map will, in probability, be very close to the ratio of the lengths of the two texts. In other words, the slope of map segments has very low variance.

## 3 The Basic Method

Omissions in translations give rise to distinctive patterns in bitext maps, as illustrated in Figure 1. The nearly horizontal part of the bitext map in region O takes up almost no part of the vertical axis. This region represents a section of the text on the horizontal axis that has no corresponding section in the text on the vertical axis — the very definition of an omission. The slope between the end points of the region is unusually low. An omission in the text on the horizontal axis would manifest itself as a nearly vertical region of the bitext map. These unusual slope conditions are the key to detecting omissions.

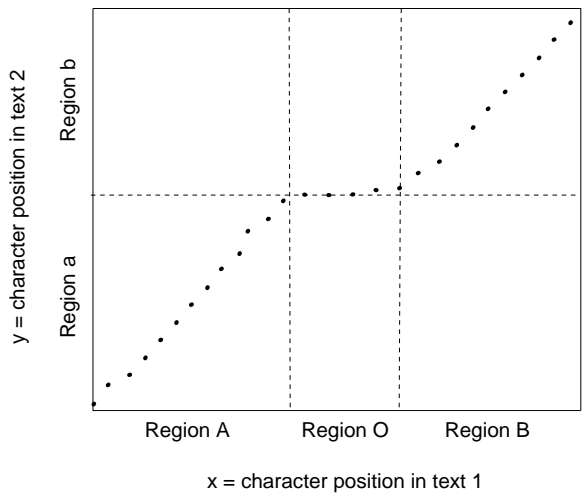

Figure 1: *An omission in bitext space. Regions A and B correspond to regions a and b, respectively. Region O has no corresponding region on the vertical axis.*

Given a noise-free bitext map, omissions are easy to detect. First, a bitext space is constructed by placing the original text on the x-axis, and the translation on the y-axis. Second, the known points of correspondence are plotted in the bitext space. Each adjacent pair of points delimits a segment of the bitext map. Any segment whose slope is unusually low is a likely omission. This notion can be made precise by specifying a slope angle threshold $t$. So, third, segments with slope angle $a < t$ are flagged as **omitted segments**.

## 4 Noise-Free Bitext Maps

The only way to ensure that a bitext map is noise-free is to construct one by hand. Simard et al. (1992) hand-aligned corresponding sentences in two excerpts of the Canadian Hansards (parliamentary debate transcripts available in English and French). For historical reasons, these bitexts are named "easy" and "hard" in the literature. The sentence-based alignments were converted to character-based alignments by noting the corresponding character positions at the end of each pair of aligned sentences. The result was two hand-constructed bitext maps. Several researchers have used these particular bitext maps as a gold standard for evaluating bitext mapping and alignment algorithms (Simard et al., 1992; Church, 1993; Dagan et al., 1993; Melamed, 1996).

Surprisingly, ADOMIT found many errors in these hand-aligned bitexts, both in the alignment and in the original translation. ADOMIT processed both halves of both bitexts using slope angle thresholds of 5°, 10°, 15°, 20°, and 25°. For each run, ADOMIT produced a list of the bitext map's segments whose slope angles were below the specified threshold $t$. The output for the French half of the "easy" bitext, with $t = 15°$, consisted of the following 10 items:

```
(26869, 29175) to (26917, 29176)
(42075, 45647) to (42179, 45648)
(44172, 47794) to (44236, 47795)
(211071, 230935) to (211379, 231007)
(211725, 231714) to (211795, 231715)
(319179, 348672) to (319207, 348673)
(436118, 479850) to (436163, 479857)
(453064, 499175) to (453116, 499176)
(504626, 556847) to (504663, 556848)
(658098, 726197) to (658225, 726198)
```

Each ordered pair is a co-ordinate in the bitext space; each pair of co-ordinates delimits one omitted segment. Examination of these 10 pairs of character ranges in the bitext revealed that

- 4 omitted segments pointed to omissions in the original translation,

- 4 omitted segments pointed to alignment errors,

- 1 omitted segment pointed to an omission which apparently caused an alignment error (*i.e.* the segment contained one of each),

- 1 omitted segment pointed to a piece of text that was accidentally repeated in the original, but only translated once.

With $t = 10°$, 9 of the 10 segments in the list still came up; 8 out of 10 remained with $t = 5°$. Similar errors were discovered in the other half of the "easy" bitext, and in the "hard" bitext, including one omission of more than 450 characters. Other segments appeared in the list for $t > 15°$. None of the other segments were outright omissions or misalignments. However, all of them corresponded to non-literal translations or paraphrases. For instance, with $t = 20°$, ADOMIT discovered an instance of "Why is the government doing this?" translated as "Pourquoi?"

The hand-aligned bitexts were also used to measure ADOMIT's recall. The human aligners marked omissions in the original translation by 1-0 alignments (Gale & Church, 1991; Isabelle, 1995). ADOMIT did not use this information; the algorithm has no notion of a line of text. However, a simple cross-check showed that ADOMIT found all of the omissions. The README file distributed with the bitexts admitted that the "human aligners weren't infallible" and predicted "probably no more than five or so" alignment errors. ADOMIT corroborated this prediction by finding exactly five alignment errors. ADOMIT's recall on both kinds of errors implies that when the ten troublesome segments were hand-corrected in the "easy" bitext, the result was very likely the world's first noise-free bitext map.

## 5  A Translators' Tool

As any translator knows, many omissions are intentional. Translations are seldom word for word. Metaphors and idioms usually cannot be translated literally; so, paraphrasing is common. Sometimes, a paraphrased translation is much shorter or much longer than the original. Segments of the bitext map that represent such translations will have slope characteristics similar to omissions, even though the translations may be perfectly valid. These cases are termed **intended omissions** to distinguish them from omission errors. To be useful, the omission detection algorithm must be able to tell the difference between intended and unintended omissions.

Fortunately, the two kinds of omissions have very different length distributions. Intended omissions are seldom longer than a few words, whereas accidental omissions are often on the order of a sentence or more. So, a simple heuristic for separating the accidental omissions from the intended omissions is to sort all the omitted segments from longest to shortest. The longer accidental omissions will float to the top of the sorted list.

Translators can search for omissions when they finish a translation, just like they might run a spelling checker. A translator can find omission errors by scanning the sorted list of omitted segments from the top, and examining the relevant regions of the bitext. Each time the list points to an accidental omission, the translator can make an appropriate correction in the translation. If the translation is reasonably complete, the accidental omissions will quickly stop appearing in the list and the correction process can stop. Only the smallest errors of omission will remain.

## 6  The Problem of Noisy Maps

The results of Section 4 demonstrate ADOMIT's potential. However, such stellar performance is only possible with a nearly perfect bitext map. Such bitext maps rarely exist outside the laboratory; today's best automatic methods for finding bitext maps are far from perfect (Church, 1993; Dagan et al., 1993; Melamed, 1996). At least two kinds of map errors can interfere with omission detection. One kind results in spurious omitted segments, while the other hides real omissions.

Figure 2 shows how erroneous points in a bitext map can be indistinguishable from omitted segments. When such errors occur in the map,

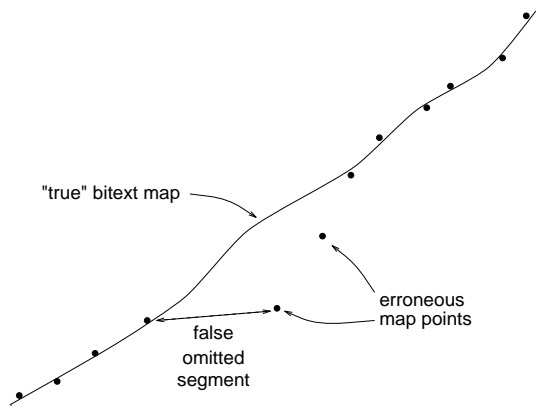

Figure 2: *An undetectable error in the bitext map. A real omission could result in the same map pattern as these erroneous points.*

ADOMIT cannot help but announce an omission where there isn't one. This kind of map error is the main obstacle to the algorithm's precision.

The other kind of map error is the main obstacle to the algorithm's recall. A typical manifestation is illustrated in Figure 1. The map segments in Region O contradict the injective property of bitext maps. Most of the points in Region O are probably noise, because they map many positions on the x-axis to just a few positions on the y-axis. Such spurious points break up large omitted segments into sequences of small ones. When the omitted segments are sorted by length for presentation to the translator, the fragmented omitted segments sink to the bottom of the list along with segments that correspond to small intended omissions. The translator is likely to stop scanning the sorted list of omissions before reaching them.

## 7  ADOMIT

ADOMIT alleviates the fragmentation problem by finding and ignoring extraneous map points. A couple of definitions help to explain the technique. Recall that omitted segments are defined with respect to a chosen slope angle threshold $t$: Any segment of the bitext map with slope angle less than $t$ is an omitted segment. An omitted segment that contains extraneous points can be characterized as a sequence of minimal omitted segments, interspersed with one or more interfering segments. A **minimal omitted segment** is an omitted segment between two adjacent points in the bitext map. A **maximal omitted segment** is an omitted segment that is not a proper subsegment of another omitted segment. **Interfering segments** are subsegments of maximal omitted segments with a slope angle *above* the chosen threshold. Interfering segments are always delimited by extraneous map points. If it were not for interfering segments, the fragmentation problem could be solved by simply concatenating adjacent minimal omitted segments. Using these definitions, the problem of reconstructing maximal omitted segments can be stated as follows: Which sequences of minimal omitted segments resulted from fragmentation of a maximal omitted segment?

A maximal omitted segment must have a slope angle below the chosen threshold $t$. So, the problem can be solved by considering each pair of minimal omitted segments, to see if the slope angle between the starting point of the first and the end point of the second is less than $t$. This brute force solution requires approximately $\frac{1}{2}n^2$ comparisons. Since a large bitext may have tens of thousands of minimal omitted segments, a faster method is desirable.

Theorem 1 suggests a fast algorithm to search for pairs of minimal omitted segments that are farthest apart, and that may have resulted from fragmentation of a maximal omitted segment. The theorem is illustrated in Figure 3. $\bar{B}$ and $\vec{T}$ are mnemonics for "bottom" and "top."

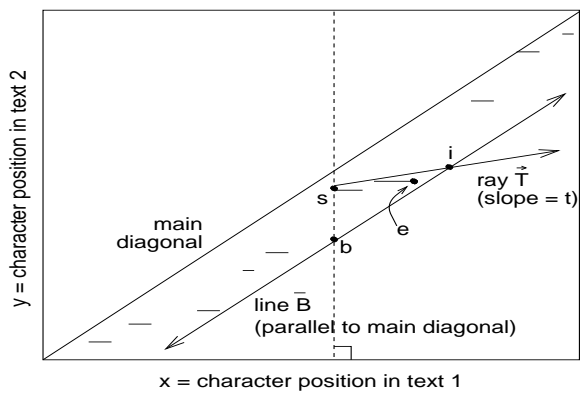

Figure 3: *An efficient search for maximal omitted segments. The array of minimal omitted segments lies above line $\bar{B}$. Any sequence of segments starting at $s$, such that the slope angle of the whole sequence is less than $t$, must end at some point $e$ in the triangle $\triangle sib$.*

**Theorem 1** *Let $A$ be the array of all minimal omitted segments, sorted by the abscissa of the left end point. Let $\bar{B}$ be a line in the bitext space, whose slope equals the slope of the main diagonal, such that all the segments in $A$ lie above $\bar{B}$. Let $s$ be the left endpoint of a segment in $A$. Let $\vec{T}$ be a ray starting at $s$ with a slope angle equal to the chosen threshold $t$. Let $i$ be the intersection of $\bar{B}$ and $\vec{T}$. Let $b$ be the point on $\bar{B}$ with the same abscissa as $s$. Now, a maximal omitted segment starting at $s$ must end at some point $e$ in the triangle $\triangle sib$.*

**Proof Sketch:** *$s$ is defined as the left end point, so $e$ must be to the right of $s$. By definition of $\bar{B}$, $e$ must be above $\bar{B}$. If $e$ were above $\vec{T}$, then the slope angle of segment $\bar{se}$ would be greater than the slope angle of $\vec{T} = t$, so $\bar{se}$ could not be an omitted segment.* □

ADOMIT exploits Theorem 1 as follows. Each minimal omitted segment z in A is considered in turn. Starting at z, ADOMIT searches the array A for the last (*i.e.* rightmost) segment whose right end point $e$ is in the triangle $\triangle sib$. Usually, this segment will be z itself, in which case the single minimal omitted segment is deemed a maximal omitted segment. When $e$ is not on the same minimal omitted segment as $s$, ADOMIT concatenates all the segments between $s$ and $e$ to form a maximal omitted segment. The search starting from segment z can stop as soon as it encounters a segment with a right end point higher than $i$. For useful values of $t$, each search will span no more than a handful of candidate end points. Processing the entire array A in this manner produces the desired set of maximal omitted segments very quickly.

## 8 Evaluation

To accurately evaluate a system for detecting omissions in translations, it is necessary to use a bitext with many omissions, whose locations are known in advance. For perfect validity, the omissions should be those of a real translator, working on a real translation, detected by a perfect proof-reader. Unfortunately, first drafts of translations that had been subjected to careful revision were not readily available. Therefore, the evaluation proceeded by simulation. The advantage of a simulation was complete control over the lengths and relative positions of omissions. This is important because the noise in a bitext map is more likely to obscure a short omission than a long one.

The simulated omissions' lengths were chosen to represent the lengths of typical sentences and paragraphs in real texts. A corpus of 61479 *Le Monde* paragraphs yielded a median French paragraph length of 553 characters. I had no corpus of French sentences, so I estimated the me-dian French sentence length less directly. A corpus of 43747 *Wall Street Journal* sentences yielded a median English sentence length of 126 characters. This number was multiplied by 1.103, the ratio of text lengths in the "easy" Hansard bitext, to yield a median French sentence length of 139. Of course, the lengths of sentences and paragraphs in other text genres will vary. The median lengths of sentences and paragraphs in this paper are 98 and 627 characters, respectively.

The placement of simulated omissions in the text was governed by the assumption that translators' errors of omission occur independently from one another. This assumption implied that it was reasonable to scatter the simulated omissions in the text using any memoryless distribution. Such a distribution simplified the experimental design, because performance on a fixed number of omissions in one text would be the same as performance on the same number of omissions scattered among multiple texts. As a result, the bitext mapping algorithm had to be run only once per parameter set, instead of separately for each of the 100 omissions in that parameter set.

A useful evaluation of any omission detection algorithm must take the human factor into account. A translator is unlikely to slog through a long series of false alarms to make sure that there are no more true omissions in the translation. Several consecutive false omissions will deter the translator from searching any further. On average, the more consecutive false omissions it takes for a translator to give up, the more true omissions they will find. Thus, recall on this tasks correlates with the amount of patience that a translator has. Translator patience is one of the independent variables in this experiment, quantified in terms of the number of consecutive false omissions that the translator will tolerate.

Separate evaluations were carried out for the Basic Method and for ADOMIT, and each method was evaluated separately on the two different omission lengths. The 2x2 design necessitated four repetitions of the following steps:

1. 100 segments of the given length were deleted from the French half of the bitext. The position of each simulated omission was randomly generated from a uniform distribution, except that, to simplify subsequent evaluation, the omissions were spaced at least 1000 characters apart.

2. A hand-constructed bitext map was used to find the segments in the English half of the bitext that corresponded to the deleted French segments. For the purposes of the simulation, these English segments served as the "true" omitted segments.

3. The SIMR bitext mapping algorithm (Melamed, 1996) was used to find a map between the original English text and the French text containing the simulated omissions.[1]

4. The bitext map resulting from Step 3 was fed into the omission detection algorithm. The flagged omitted segments were sorted in order of decreasing length.

5. Each omitted segment in the output from Step 4 was compared to the list of true omitted segments from Step 2. If any of the true omitted segments overlapped the flagged omitted segment, the "true omissions" counter was incremented. Otherwise, the "false omissions" counter was incremented. An example of the resulting pattern of increments is shown in Figure 4.

6. The pattern of increments was further analyzed to find the first point at which the "false omissions" counter was incremented 3 times in a row. The value of the "true omissions" counter at that point represented the recall achieved by translators who give up after 3 consecutive false omissions. To measure the recall that would be achieved by more patient translators, the "true omissions" counter was also recorded at the first occurrence of 4 and 5 consecutive false omissions.

7. Steps 1 to 6 were repeated 10 times, in order to measure 95% confidence intervals.

The low slope angle thresholds used in Section 4 are suboptimal in the presence of map noise, because much of the noise results in segments of very low slope. The optimum value $t = 37°$ was determined using a separate development bitext. With $t$ frozen at the optimum value, recall was measured on the corrected "easy" bitext.

Figures 5 and 6 plot the mean recall scores for translators with different degrees of patience. ADOMIT outperformed the Basic Method by up to 48 percentage points. ADOMIT is also more robust, as indicated by its narrower confidence intervals. Figure 6 shows that ADOMIT can help translators catch more than 90% of all paragraph-size omissions, and more than one half of all sentence-size omissions.

ADOMIT is only limited by the quality of the input bitext map. The severity of this limitation is yet to be determined. This paper evaluated ADOMIT on a pair of languages for which SIMR can reliably produce good bitext maps (Melamed, 1996). SIMR will soon be tested on other language pairs. ADOMIT will become even more useful as better bitext mapping technology becomes available.

---

[1] SIMR can be used with or without a translation lexicon. Use of a translation lexicon results in more accurate bitext maps, which make omission detection easier. However, wide-coverage translation lexicons are rarely available. To make the evaluation more representative, SIMR was run without this resource.

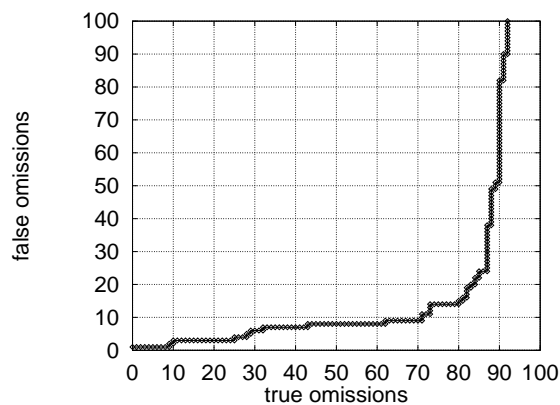

Figure 4: *An example of the order of "true" and "false" omissions when sorted by length. Horizontal runs correspond to consecutive "true" omissions in the output; vertical runs correspond to consecutive "false" omissions. In this example, the first run of more than 3 "false" omissions occurs only after 87 "true" omissions.*

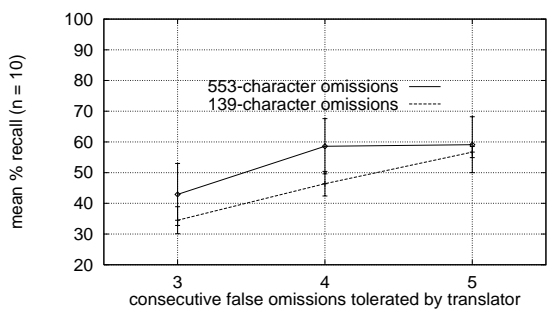

Figure 5: *Mean Basic Method recall scores with 95% confidence intervals for simulated translators with varying degrees of patience.*

## 9 Conclusion

ADOMIT is the first published automatic method for detecting omissions in translations. ADOMIT's performance is limited only by the accuracy of the input bitext map. Given an accurate bitext map, ADOMIT can reliably detect even the smallest errors of omission. Even with today's poor bitext mapping technology, ADOMIT finds a large enough proportion of typical omissions to be of great practical benefit. The technique is easy to implement and easy to integrate into a translator's routine. ADOMIT is a valuable quality control tool for translators and translation bureaus.

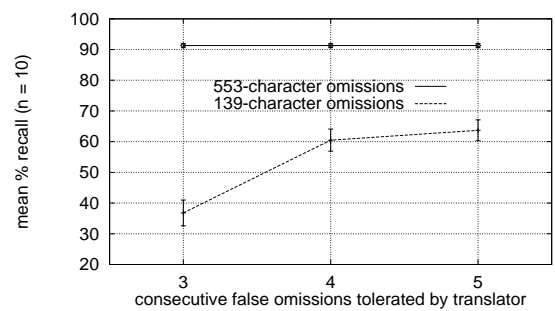

Figure 6: *Mean ADOMIT recall scores with 95% confidence intervals for simulated translators with varying degrees of patience.*

## Acknowledgements

This research began while I was a visitor at the Centre d'Innovation en Technologies de l'Information in Laval, Canada. The problem of omission detection was suggested to me by Elliott Macklovitch. I am grateful to the following people for commenting on earlier drafts: Pierre Isabelle, Mickey Chandrasekar, Mike Collins, Mitch Marcus, Adwait Ratnaparkhi, B. Srinivas, and two anonymous reviewers. My work was partially funded by ARO grant DAAL03-89-C0031 PRIME and by ARPA grants N00014-90-J-1863 and N6600194C 6043.